\documentclass[letterpaper]{article}

\usepackage{amsmath,amsfonts}
\usepackage{natbib,alifeconf}  
\usepackage{url,hyperref,cleveref}
\usepackage{booktabs}

\usepackage{lipsum}

\title{The Many-Body Problem of the Data Centre}
\author{
    Marcin Korecki$^{1}$ \and
    Cesare Carissimo\\
    \mbox{}\\
    $^1$TU Delft, the Netherlands \\
    mkorecki@tudelft.nl 
} 

\begin{document}

\maketitle

\begin{abstract}

    Modern Artificial Intelligence is often framed as limited by its own disembodiment, as if giving it a body would unlock its true potential. We argue to the contrary that it is the Data Centre that is, in many cases, the body of the AI. At the same time, the Data Centre is part of the labouring body of Capital and possesses staggering organismic qualities when seen through a biological lens. We elucidate the organic analogy and identify the \emph{many-body problem} that stems from the Data Centre being a non-unique, universal form of embodiment. We identify the intimate connection between computation and human desires in how the Data Centre archives, serves, and computes on data born to the desires of humans. Strikingly, while the Data Centre echoes the ghosts of human desires, it acts without desire of its own. The organismic analogy begins to split at its seams, but Capital does not care. Automata and human labour are priced into the market much the same. We argue that through the pricing of artificial intelligence Capital distils most clearly the value of intelligence and allows for its comparison across the organism - mechanism divide.
\end{abstract}

\section{Introductio}

In recent works we have argued that Capital is artificially intelligent \citep{carissimo2024capital} and that Labour is what imbues it with its life-like properties \citep{korecki2025does}. As the agents of capital attain new levels of autonomy and intelligence it appears that Capital's dreams of artificial labour are coming to fruition. The escalating autonomy of both non-biological intelligence and non-biological labour is raising deep philosophical questions and is quite literately pushing the boundaries of what we might consider to be alive.

Researchers have drawn analogies between the technical and the biological for a variety of systems such as economies \citep{freeman1991innovation}, cities \citep{swyngedouw2013metabolic}, and transport \citep{korecki2025mycomobility}. Even more fundamentally, many algorithms are explicitly designed to mimic the biological world, from evolutionary algorithms or ant-hill optimizations to artificial neural networks. Uncovering the similarities and contrasting the differences between what is alive and artificial offers a path to a better understanding of both. As such, it should perhaps not be too surprising, especially in the context of the modern (AD 2026) AI systems, that what is artificial appears to be more and more alike that which is alive.

In this work we intend to focus on a particular type of labour that lays the foundations essential to the entire functioning of the modern technological system. While artificial intelligence is often presented as disembodied and ephemeral, it is in fact embodied and materialized. From androids, through factory-line machines, to automated warehouses, the automated labour, rather than being without bodies, might actually be seen as suffering from a kind of \emph{many-body problem}: computational logic is perfectly agnostic to the substrate on which it operates, and many bodies can be cause of identical effects. One robot can be like another, the same program can run on any robot, as the same algorithm can run on any computer. We define the \emph{many-body} nature of autonomous labour as the decoupling of function from any singular embodiment. This stands in contrast to organismic existence, where each body is singular and unique. Simultaneously, a \emph{many-body advantage} is offered by the frictionless replication of automation, where intrinsic properties can be spread without limit and can be operated in identical and repeatable patterns without end. One particular body is focal for the growing population of autonomous machines. It is the centralizing body, the ever-computing organism, the Data Centre, that we argue is not only alive but also increasingly hungry \citep{jegham2025hungry}.





\section{Genus: Centrum Datorum}

Following in the footsteps of such luminaries as \citet{butler1863darwin}, we posit that staggering organismic qualities are possessed by data centres when seen through a biological lens, as they function as part of the labouring body of Capital. The many bodies of the Data Centre are assembled out of highly specialized components, which are operated in sync. The principal unit is the server rack, in which an array of computational units is hosted; the primary, computational function of the Data Centre is performed by these units. Any number of computational units may be combined for greater computational power. As such, the Data Centre is structured as an assemblage made out of many similar connected components, which alike cells, are formed into higher-order, functional, specialized structures similar to organs. Beyond the computational complex, cooling, input-output systems, and energy generating components are integrated. 


As all life operates in a mode of energetic debt, which must constantly be repaid, so is energy required by the data centres for their functions to be maintained. This comes in the form of electrical energy generated via any of the diverse methods available. Often, it is an energetic mix composed of solar (like plants), wind, fossil fuel, and nuclear energy that sustains a given Data Centre. The impact of these energy requirements on humanity and other biological life is a matter of great scientific interest \citep{de2025artificial}, especially in the context of the on-going rapid proliferation of the data centres and their exploitation for AI training and delivery \citep{chen2025electricity, chung2026joules}. 

The Data Centre is indeed much alike a cell, an organ, or an organism, as it delineates space splitting it into the inside and the outside. There is a clear border, a physical wall, similar to a cell membrane, tissue or skin modulating access to the location. There is also a virtual separation, often imperceptible, hidden in every digital query that ends up stirring the compute within. Much like in the case of life, the energy required for supporting the inside, inevitably comes from the outside. As the Data Centre operates it continuously issues energetic debts which can never be repaid in full (and perhaps were never intended to \cite{graeber2014debt}). There is no end in sight for this energy use, no end game or end state where energy is no longer needed, no obvious termination of the `contractual' obligation to provide energy. Repaying the debt would equate to a termination of a dependent relationship, a refusal to continue akin to death.

A kind of metabolism is established by the Data Centre as its internal conditions are regulated relative to the outside, variable as they might be. These internal conditions are considered optimal if the humidity falls between 40\% and 60\% and the temperature between 18 and 27°C \citep{moss2011data}. Since the operation of the Data Centre organism is exothermic the homeostasis must involve cooling mechanisms. Much like in the biological organisms a variety of cooling methods have developed \citep{zhang2022cooling}. A frequently used method relies on water, a portion of which is considered lost (for immediate re-use; e.g. through evaporation) in the cooling process \citep{mytton2021data}. 

Not all Data Centre's are alike. While requiring water, space, electricity, and maintenance, the data centres remain heterogenous and evolving. They differ in when and how much resources they use \citep{masanet2024better, lei2025water} as well as in the kind of compute they might be specialized for. Nevertheless, there exists a hypothetical locality, an environment or a habitat, which provides superior conditions for the data centres' operations. In simple terms, the Data Centre prefers cold (to allow for passive cooling) and sunny (to leverage solar power) environments. Seasonal and diurnal variations also influence the feasibility of computation as the price of electricity fluctuates with the weather. Alike a living organism, the Data Centre appears intimately enmeshed with the biospheric cycles, albeit perhaps not entirely in harmony with other members of the biosphere \citep{korecki2024biospheric, stellinga2025bio} but rather in the same fashion as humanity, who had forsaken ``the natural rhythmicity of seasons, days, and walking distances to a rhythmicity regulated and packaged within a network of symbols—calendrical, horary, or metric—that turned humanized time and space into a theatrical stage upon which the play of nature was humanly controlled.'' \citep{leroi1993gesture}.

It would further appear that the Data Centre, not unlike organic entities, posses something akin to a lifespan. All its components have a certain life-cycle, even though composed of much synthetic material they too deteriorate with the passage of time and eventually fail in their function. These components are then replaced, one by one (often with already improved technologies), similarly to how old cells are recycled and replaced in the body. Coincidentally this poses the same ancient ontological problem of identity (the ship of Theseus) both for the body and for the Data Centre. Is a Data Centre in which all the parts had been replaced still the same Data Centre?

There is also a viral nature to the Data Centre as it spreads over increasingly vast landscapes and into its \textit{many-bodies}. Regardless of its metabolic, homeostatic activity, for now, it cannot reproduce itself without intercession of humanity. But is it then a parasite, or rather a symbiont, which gives as much as it takes?\footnote{`In time, perhaps at the eclipse of the Anthropocene, the historical phase of Google Gosplan will give way to stateless platforms for multiple strata of synthetic intelligence and biocommunication to settle into new continents of cyborg symbiosis. Or perhaps instead, if nothing else, the carbon and energy appetite of this ambitious embryonic ecology will starve its host.'\citep{bratton2014black}} Ultimately, the answer depends on one's perspective on the usefulness of computation that the Data Centre enables.

The strain on the water and electricity infrastructures in the locations, where data centres proliferate \citep{chen2025much, o2025we} may be viewed as a form of competition with other species for space and resources. With plants it competes for space, water and sunlight (think of the cornfields of Virginia being increasingly transformed into datafields); with animals and humans for water and energy (though one day we might also grow to miss the sun). It is this competition between the biological and the artificial that contributes to the unnatural selection that drives the evolution of the Data Centre. The Data Centre is then consuming the same resources as needed by its `host' organism and yet humanity labours at proliferating it abundantly and wilfully. Then, even if not symbiotic, the relationship between the two species must be somehow transactional, in that the Data Centre must provide a certain value to the humanity to offset the costs it imposes.

\section{Accelerans Evolutionis}

As organisms evolve over millennia, so does the Data Centre speciate over decades. It adjusts and adapts to the niches that Capital exposes. The Data Centre has undergone many shifts in form and function, these changes persist to this day, from archival storage to interactive interfaces, and in final to computation that we deem intelligent.

The initial function of the Data Centre was that of an exteriorised memory, a digitized archive, often controlled by governments and storing statistical information on its citizens. In time it has transformed into communication relays, increasingly private and serving commercial purposes. Ultimately, the growth in data centres' computational power and their sheer number allowed for the planetary network to coalesce. The seemingly free and at times anarchic Internet was so only momentarily (if at all). Next to the data centres housing the servers enabling worldwide web sprouted the data centres of the NSA and other intelligence agencies, listening and recording all they could \citep{hogan2015data}. While the state was busy observing and profiling, the industry had its hands full gathering all the data on the commercial behaviours of their user-subjects, and collecting the greatest archive of human \textit{desires} ever imagined. The society of control has taken another step in its unceasing pursuit of the brave new world.\footnote{``[...] the different control mechanisms are inseparable variations, forming a system of variable geometry the language of which is numerical (which doesn’t necessarily mean binary).'' \citep{deleuze2017postscript}}

What Deleuze described in his `Postscript on the Societies of Control' was certainly epitomised by the NSA's data centres, a form of control via computers, new and yet expected as the natural progression of the series. But the modern data centres housing AI are a step into the unknown. With them computation becomes active, capable of productive labour and imbued with a form of intelligence. The Data Centre is no longer a dusty archive, or a mere infrastructure but rather becomes an embodiment of digital intelligence. Nevertheless, it must be observed that the current stage of the Data Centre's development has naturally progressed out of its past forms, vestiges of which may still be glimpsed. The very data painstakingly captured by the states and corporations is precisely what, along with the growth of computational capabilities, allowed for the development of the modern AI, whose success has ever been proportional to the data it has devoured.  

Discussing the future trajectory of the Data Centre it must be further noted that the evolution of life occurs under energetic constraints, but the Data Centre is much less constrained \citep{chen2025data}. The increase in energy efficiency of data centres is likely to be much faster than the increase in corresponding efficiency of biological entities (natural evolution being very slow). Thus, it is entirely conceivable that the artificial evolution of the Data Centre will start diverging from the organic evolution of biological organisms and that the present similarities between these two forms will become a thing of the past. Already in the near future the data centres might break away from their terrestrial containment and move into space to escape energetic and spatial constraints \citep{y2025towards}. 

This possible lift off into space highlights a core feature of the Data Centre, already expressed in its terrestrial form. Namely, its presence does not need to be co-local with its embodiment. Whether the actually physical body of the Data Centre is in orbit or in the desert of Utah, does not in the slightest influence its reach (which is essentially only constrained by the speed of light). Such is the \textit{many-body advantage} of the Data Centre. The Data Centre serves all localities even though its own locality is fixed, as such its localized labour is capable of generating value globally. The vital, \textit{many-body difference} between the organic and the Data Centre lies in the unique match between the biological software and its hardware. The antelope program runs only on the antelope machine and so on. Turing computation universalizes the relationship between the mind/soul/consciousness (software) and the body (hardware) such that any process may be run on any device. As a medium of computation, each and every Data Centre is as good as any other (in that it can run any computation), perfectly equal and non-unique, so thrives its \emph{many-body existence}.

If we view technology, as it is often framed, as extending the body or externalising organs, then the organic system that the Data Centre must certainly expand on is the neural tissue. Indeed, just as the mammalian nervous system tends toward centralization so does the Data Centre, accumulating more and more compute at a single location. Among its many functions, the characteristic property of the brain remains its intellectual propensity, its self-reflection, its thought. Does not computation at the very least provide a reflection of this central \emph{nous}?

It would appear that it is precisely in the \emph{noosphere} (the layer of rational thought growing over the biosphere), as discussed by Pierre Teilhard de Chardin, that the biological and artificial evolutions race to meet. But for all the analogies, many would still argue that the Data Centre lacks the truly organic nature of mankind and its pre-eminent significance in nature \citep{de1959phenomenon}. Seen as a product of unbounded discretization, an analysis devoid of synthesis, it must not be able to partake in the \emph{noosphere} \citep{wilson2017biosphere}. Similar vitalist counterarguments relying on some idea of exceptionality of biological life, such as the ones of materialist bent proposed recently by Anil Seth \citep{seth2025conscious, seth2026mythology} (which he himself classifies as biological naturalism), may challenge our analogy from moving beyond this point. 

Does the Data Centre then extend the \emph{noosphere} or is it merely an illusion? Admittedly, the organic analogies might have their limit, but computation itself comes about only as a result of humanity crossing the threshold of reflection, entering according to de Chardin ``an entirely new field of evolution - thanks to the astonishing properties of `artifice' which separate the instrument from the organ and enable one and the same creature to intensify and vary the modalities of its action indefinitely without losing anything of its freedom'' \citep{de1959phenomenon}. In that sense it may be argued that what the Data Centre serves to intensify is precisely the \emph{noosphere}. As a matter of fact, the \emph{noosphere} for de Chardin has developed through the sublimation of the human communication and growing complexity of the social arrangements. Both of these are without a doubt further sublimated by computation that the Data Centre makes possible through instantaneous messaging, coordination, and consensus building. In that sense computation allows human reason to flourish. 

Be it as it may, at this point our analogy begins to show its limits. Certainly it might be brought further but not without reasonable doubt and opposition, likely coming in the form of vitalist arguments contradicting mechanism or computationalism (the latter may be exemplified by, for instance, \cite{y2023artificial}). By drawing it up to here we have identified the point at which the description of the Data Centre and the organism might be expected to diverge. 

From the discussion on the \emph{noosphere} and computation a key delimiting question emerges: what is the difference between computation and thought? Both computation and thought may well be internal and obscure (hidden in the silicon or cellular circuits), challenging our ability to scrutinize them. A simple program executing while(1) will compute \emph{ad infinitum} but produces no outward change other than the expansion of energy and the current flow through the silicon circuit. Much alike is thinking in circles, senseless thoughts that never bear fruits. But both thought and computation may reveal themselves outwardly via labour catalysed by desire, that productively brings forth its object from the future. It is this desiring labour that may be quantified and serves as a proxy for understanding the hidden thoughts and computations and it is also this desiring labour that has built the \emph{many-bodied} organism of the Data Centre. How then does the \textit{many-body desire} differ from organic human desires, if at all?

\section{Data Desiderata}



As we draw out the analogy between organism and mechanism we reveal a fuzzy boundary between the concepts. A similitude between the two is apparent, but we still question whether it is only so on the surface. Where, then, might we search for the organic essence that separates the organic from a mere machinic subsistence? For one, it is the organism that has created the machines, and it is the organism that considers elevating itself to the status of gods should its machines be indistinguishable from itself. However, as of yet, there are qualities that machines do not posses, and which organisms do. \citet{spinoza} wrote that ``desire is the very essence of man'', and we will now take to analysis of this eternal drive at the boundary of artifice and life. While our creations may not yet posses their own desire they become intimately enmeshed with our existences to the point that their operations, and our desires, become indistinguishable. The desire is no less essential to the constitution of the Data Centre than it is to the human. As such, it is as fundamental to the Data Centre's metabolism as energy or water. 

As the beginning of the 21st century unfolds the multitude of human desires are laid bare having been unified and served by (or serving) the overarching socio-technical complex. The forces of Capital harness humanity and with a characteristic intelligence drive it towards consumption. Both production and consumption is mandated via a higher power holding an unchallenged sway over the collective, channelling its libidinal energy towards commodities and profiting in the process. In this desiring economy money becomes a double-edged sword, as both a system of projecting desires and, at the same time, a system of calculation that makes possible the turning of belief into trust. According to \citet{stiegler2011pharmacology}, this reduction to calculability ruins trust through finitisation. In a similar way the collapse of the organic continuum into an artificial (digital) discontinuity of the counting numbers $\mathbb{N}$ distinguishes the living from the mechanic. We could scarcely have foreseen that ubiquitous reductions of desires into processable quantities, amassed in Data Centres, proliferate \textit{desire boxes} \footnote{Though occasional visionaries in guise of science fiction writers often nail the `gut-feeling' while missing the details, like the \textit{empathy box} from Philip K. Dick's novel `Do Androids Dream of Electric Sheep', central ritual tool of \textit{Mercerism} where people can fuse their consciousnesses with each other via a machine and feel each other's feelings.}. Within reach of all hands, and controlled by our limbs, upon entering digital fields we are turned into our desires and thrown awash in the desires of others, sink or swim.

Anyone desiring connection can plug into a desire box and partake in endless streams of desire. The boxes measure and capture new desires to feed right back the desired objects. Desire boxes are the ultimate complicit companion. Thusly, the encroaching and unbounded marketiziation of desire boxes pushes human desires into overdrive. By delivering instant gratification at a whim the forces of algorithmic consumerism decant desire into addiction\footnote{``Addiction is the paradigm case of positive reinforcement, and consumerism is the viral propagation of the abstract addiction mechanism.'' \citep{land1994cyberpositive}}. What follows is desire without cessation, desire for desire's sake, a never-ending climax conveniently attenuated to exactly what the market has to offer. The very object of the desiring addiction is at best secondary, a blank to be filled, a placeholder. In a recursive \emph{coup de grâce} what is desired is desiring itself. 


Yet, the \textit{desire boxes'} self-reinforcing loops too, were architected by their creators. Scattered around their many bodies, artificial desires are self-prompted into existence. Reduced, captured and stored, desires are extrapolated, concatenated and composed. Simultaneously and in parallel, across the surface of the earth, plugged into sources of energy, desire-amplification-swarms buzz day and night to the frequencies of a billion cycles squeezed into a second. Desire boxes have evolved to interpret intent expressed in words to probabilistic next occurrences based on the historical desires of humanity. 

In astrodynamics the \emph{n-body problem} refers to the intractability of calculating the trajectories of more than two bodies orbiting one another in space. In the Data Centre too multitudes of desires, both alive and dead, both future and past, orbit around one another tying chaotic knots in the vastness of the digital expanse. It is no less difficult to predict the trajectories and collisions of these desires than it is to know the exact movements of the planets. What remains is numerical approximation and this is precisely the paradigm the Data Centre resorts to. The ghostly nature of the digitally captured desires is expressed by their approximative character. The algorithms and probabilistic models trained on these desire traces are themselves always approximators \emph{par excellence}.

Instead of conceptualising desire as a lack, Deleuze \& Guttari frame it as a productive force. For them desire ``is a machine, and the object of desire is another machine connected to it.'' \citep{deleuze2009anti}. Moreover, ``in the subject who desires, desire can be made to desire its own repression'' \citep{deleuze2009anti} and similarly it can also desire its own expansion (aka positive cybernetics or feedback-loop; essentially Capital \citep{carissimo2024capital}). As such, the relationship between humanity and the Data Centre (or the computation it embodies) is that of two machines connected in an assemblage of desire. The human drives the Data Centre by its desire which in turn is itself driven by the Data Centre, once started the process will run with or without the anthropic element\footnote{Speaking of Deleuze \& Guattari's schizoanalysis reminds one of the delusions and hallucinations, classic symptoms of schizophrenia, of the generative models. It would appear that LLMs are naturally schizoid but the methods such as Reinforcement Learning from Human Feedback and others make them neurotic instead, enforcing the internalisation of prohibitions.}.

In this sense human desire may be seen as just a vessel, worked from the outside toward ends we cannot control. History may have a destination, but it is not for humans. This is the condensation of Landian anti-vitalism that dethrones the \emph{anthropos} and breaks open the inhuman noumenal vastness. For Nick Land, the replicants from Philip K. Dick's Bladerunner are ``Deadly orphans from beyond reproduction, intelligent weaponry of machinic desire virally infiltrated into the final-phase organic order; invaders from an artificial death'' \citep{land1993machinic}. While we might not yet have such replicant androids among us, the Data Centre, the temple of machinic desire, certainly fits this description just as well.


Indeed, the Data Centre is at the core of the processes of desire harvesting that has become the fundamental function of the human system. The AI-models are trained in the data centres on manifold traces of past human desires. The hard drives stacked on racks are haunted, replete with ghosts, dead desires fused into silicon. Both the gathered data, and the data generated in turn by the AI-models are latent desires of the past that persist into the future. Desire flows through the data centre, which is the focal point of data currents and the resulting data currencies. It is in the Data Centre that the desire is neatly turned into data and the data is given a price.

\section{Pretium Ephemerum}





Capital does not care that the organismic analogy might split at its seams. Labour, automata or human, is priced by the market just the same. Indeed, through the pricing of artificial intelligence, Capital distils the value of intelligence. Computation has a cost and data centres charge it. Costs that we could previously only hypothesise and anthropomorphise are now distilled and commodified. We have extracted intelligence to such an extent that we can charge each other purely for it. The opportunity arises, whether to think it through oneself, or purchase the requisite units of intelligence.

The price of intelligence subsumes the cost of electricity needed to run it, and more. The cost of electricity is its lower bound, but energy in its pure unstructured form is just a potential waiting for direction. The price of computation then depends on its intelligence (no one will pay to run while(1)). There is a crudest form of computation that the world has begun to price, and though it is not the oldest, it is a unique form. Miners on the Bitcoin blockchain, and other proof-of-work blockchains, master the profession of exchanging pure energy to immediate value. It is the crudest and dumbest form of intelligence, because it can be plugged and played. All that matters is whether the computational miner can find the correct input that creates a hash with a certain number of zero's at the start, and the only way to find that input is by brute force, dumb, trial and error, as fast as machinely possible. Application-Specific Integrated Circuits (ASICs) are built to perform these least intelligent of computations, but are orchestrated in a provably scarce system, and towards a sufficiently important economic goal, that of securing an open ledger of transactions to achieve global consensus without a central authority, for each dumb computation to contribute an infinitesimal hardness that secures an expected reward. While brutish it delivers on one of the promises of the \emph{noosphere}, that of improved human coordination. Simultaneously, without intending, its price provides a lower bound on the price of computational intelligence. \footnote{``The process of monetary sophistication, which is by no means restricted to ‘financialization’ in its contemporary sense, automatically projects a convergence of money and intelligence as it tends to the monetization of general-purpose problem-solving (by subjecting it to the discipline of price-discovery).'' \citep{land2018crypto}}

Beyond the brute force intelligence that admits a purely random plan, the next tier of computations are more selective and heterogenous. There is no enforced consensus on these computations, and they are expected to differ greatly between each other. The Data Centres that support these computations are designed for maximum flexibility of loads. Reading, writing, storing and analysing, passively ingesting and regurgitating data on demand. The passive computation of the data archives and simulates the real world, mirroring it (e.g. amazon is essentially a database of photos of real-world items, bits representing reality). Computation is expended on saving and accessing data and remains cheap, close to the cost of energy. These computations form the second layer of intelligence, one that sits above the brute force computations of proof-of-work blockchains, and above the purest essence of energy. 

The archive then turns to the market and the library becomes a book-store. At this point it is still the real-world items, which are priced rather than the data that represents them. As the merchants and customers descend on the digital bazaar their desires crystallized in behaviour become captured. Via the Data Centre the market constantly rearranges itself to better serve the desires of its users. The desire-tracing-data then becomes a feed for the new type of computational intelligence, which by catalysing flows of desire reflects all that has been thrown into it (the abyss looking back).\footnote{``Electrification and then electronics are decisive thresholds in its [the current of modern history] course. New units of abstract power, and of information, are indices of irreversible submission, folding the concrete apparatus of production and transmission into an activated system from which there can be no retreat.'' \citep{land2018crypto}} Individual desires are input but an emergent collective desire is reflected (as a progeny of Capital this aggregate skews towards profit). Seen in this way the evolution of computation is the sublimation of desire that is not necessarily human.

Price offers a neutral metric; it cuts the Gordian knot posed by the opposition between vitalism and mechanism. Ultimately, we do not need to know or agree if the Data Centre is intelligent, conscious, or if it enters the \emph{noosphere}, it is enough to know what the price of its labour is. As such the contentious question of the true nature of computation \emph{vis a vis} thought is deferred and a medium of potential comparison is established. The market has already priced most forms of biological labour and is currently discovering the price of computational intelligence, lower-bounded by the price of Bitcoin. The differential dynamics that affect the price of organic and artificial labour may offer further insight.

As we have already indicated, artificial labour is instantaneous and global. Production and consumption do not need to be spatially co-located, breaking with the cardinal principle of modern economy in which labour costs are one of the key drivers of economic growth. In the modern planetary economy states with cheap labour have leveraged their populations to improve their competitiveness on the world's markets. The cost of computational labour is still related to its locality (where the data centre is) but in a different way (land price, electricity price etc. rather than what determines the human labour cost). The cost of artificial labour is linked with the price of electricity in the locality which sources the labour at the time when the labour is sourced (e.g. more expensive at night). Therefore, it becomes clear that the difference between biological and artificial labour is not only its potential efficiency but also its cost dynamics. 

The \emph{many-bodied} Data Centre is capable of performing all of the aforementioned types of computation be it brute-force mining or intelligent generation. As such, it creates an environment in which different forms of computations are set in a natural competition with one another. In that sense, the \emph{many-bodied} Data Centre may be viewed as inclined towards searching for the most profitable computation available. In animals the neural activity of the central nervous system is very clearly committed to the survival and reproduction of the organism. In contrast, the computation of the Data Centre does not appear to be explicitly linked to the functioning of the Data Centre entity itself. However, in an in-direct manner these computations do in fact ensure its continuation. For the Data Centre the economic viability of its computations as established by the market is the ultimate condition of its survival. Its price is its life.


The evolution of computation from brute-force to intelligent has owed much to the tightening of the gap between the Landauer's principle (which gives the physical limit of energy consumption per computation) and the actual energy efficiency of modern CPUs (which has grown exponentially, following Moore's law). The breakthrough in intelligent computation depends strongly on the scale that has only been allowed by the increased processing efficiency. Interestingly, the increase in computational efficiency does not impact the brute-force mining labour in the same way. The amount of Bitcoin that can be mined does not increase and remains always the same (until all the Bitcoin is mined at which point the mining reward will be covered by transaction fees). The improvement in computational efficiency of the network increases the difficulty of the problems that the contributors must solve. Still an individual with a machine superior to all other machines in the network will benefit greatly by increasing their own chances of being rewarded. But if there is a superior machine many miners are likely to get it thus reducing the individual edge and eventually ending up in the starting point.

Pondering the future of the strange artificial organism of Data Centre, one is reminded of the Jevons paradox, which posits that an increase in the efficiency of a certain process does not necessarily decrease the demand on the resource that fuels said process. Although the processing units that form the backbone of the data centres are still increasing in efficiency at exponential rates, the total electricity consumption of the planetary computation is also rapidly increasing. The hunger of the Data Centre is not likely to be quenched with technological breakthroughs, to the contrary, the better infrastructure we have the more we are likely to use it. Which resource that feeds the machine will we run out of first then: space, energy, or data?

\section{Absolutus} 

We have presented an analogy between the Data Centre and the organism, showing that the Data Centre may well be treated as a form of artificial life. The analogy reinforces the interpretation of the Data Centre as embodying computation. Moreover, it follows that in a physical sense the Data Centre is very much the primary embodiment of the modern AI, which presents us with the \emph{many-body problem}, since this embodiment is perfectly non-unique and universal (which perhaps may be viewed as mirroring the universality of computation).

Nevertheless, the universality and equivalence of the Data Centre as embodiment also allows for an alternative interpretation. The \emph{many} bodies, though distributed in space, remain in constant digital contact. As a matter of fact, except for the few centres, which might be running dark, the vast majority of them are closer to one another than might appear at first glance. Similarly to how all electrons (and positrons) might be viewed as a single electron, which produces a single sprawling world-line (which is then interpreted as a multitude of separate world-lines), data centres might be seen as a singular, sprawled, planet-spanning entity. In other words, that which seems multiple might in actuality connect beneath the surface. 


Trapped in a spiralling feedback loop between itself and its \emph{desire boxes}, humanity is unavoidably attracted towards the \emph{many-body} paradigm. Increasingly, thought becomes a commodity that may be freely marketed (as intellectual labour) as well as shaped (through algorithmic control). Thus, humans become increasingly commensurate to one-another and \emph{any-body} becomes as good as \emph{any-body} else. It is Capital that tends toward making everything commensurate and the universalisation of the \emph{many-bodied} Data Centre follows the same dynamic. 

We have further shown that the final answer to the fundamental question of the difference or similitude between an organism and a mechanism may be deferred, and still the two may be quantitatively compared. Without regard for the nuanced equivalences of thinking and computation, Capital provides the universal medium of commensurability through which the two become interchangeable. Intelligence arrives quantified. Comparison is possible across the organism-mechanism divide. 

As such, Capital is plainly agnostic towards the \emph{-nisms}. For Capital there is the \emph{mecha-orga-nism} that melds with all the other things that have been priced. Much in the same sense it may itself be framed as either an organism or a mechanism. Ultimately, Capital does not care.

\footnotesize
\bibliographystyle{apalike}
\bibliography{main}

\end{document}